\documentclass[10pt]{article}
\usepackage[english]{babel}										
\usepackage[utf8]{inputenc}										
\usepackage[T1]{fontenc}										
\usepackage{amsmath,amsfonts,amssymb,amsthm,cancel,siunitx,
calculator,calc,mathtools,empheq,latexsym}
\usepackage{algorithm}
\usepackage{mathtools}
\usepackage[noend]{algpseudocode}
\usepackage{subfig,epsfig,tikz,float}		            
\usepackage{booktabs,multicol,multirow,tabularx,array}          
\usepackage{natbib}
\title{\normalsize\bf%
\uppercase{Epigenetics Algorithms: Self-Reinforcement-Attention mechanism to regulate chromosomes expression}
}
\author{%
Mohamed Djallel DILMI$^{1}$ \ , \ Hanene AZZAG$^{2}$ \ , \ Mustapha LEBBAH$^{3}$
}

\begin{document}

\date{}

\maketitle

\vspace{-0.5cm}

\begin{center}
{\footnotesize 
$^1$LIPN (CNRS UMR 7030), Université Sorbonne Paris Nord \\
$^2$LIPN (CNRS UMR 7030), Université Sorbonne Paris Nord \\
$^3$DAVID, Université Paris-Saclay \\
E-mails: dilmi@lipn.univ-paris13.fr / azzag@lipn.univ-paris13.fr / lebbah@uvsq.fr 
}
\end{center}

\bigskip
\noindent
{\small{\bf ABSTRACT.}
  Genetic algorithms are a well-known example of bio-inspired heuristic methods. They mimic natural selection by modeling several operators such as mutation, crossover, and selection. Recent discoveries about Epigenetics regulation processes that occur "on top of" or "in addition to" the genetic basis for inheritance involve changes that affect and improve gene expression. They raise the question of improving genetic algorithms (GAs) by modeling epigenetics operators. This paper proposes a new epigenetics algorithm that mimics the epigenetics phenomenon known as DNA methylation. 
  The novelty of our epigenetics algorithms lies primarily in taking advantage of attention mechanisms and deep learning, which fits well with the genes enhancing/silencing concept. 
  The paper develops theoretical arguments and presents empirical studies to exhibit the capability of the proposed epigenetics algorithms to solve more complex problems efficiently than has been possible with simple GAs; for example, facing two Non-convex (multi-peaks) optimization problems as presented in this paper, the proposed epigenetics algorithm provides good performances and shows an excellent ability to overcome the lack of local optimum and thus find the global optimum.
}


\section{Introduction}
In biology, the study of epigenetics is a very active field of research since it allows us to answer questions that genetics cannot answer. It combines the ideas of Charles Darwin and Jean-Baptiste Lamarck. The literature that allows a deep understanding of epigenetics is consequently very large. For example, \cite{Jablonka2009TransgenerationalEI} have written a comprehensive review on the mechanisms of cellular epigenetics inheritance, important concepts that we will try to mimic in our paper. 

In computer science, some research has been done in order to add epigenetics mechanisms to evolutionary computations for optimization purposes. Section \ref{Related work} presents a review of some existing papers.
In this paper, we propose a new approach to epigenetics algorithms (EGAs) by incorporating a novel attention mechanism that mimics the process of DNA methylation \cite{Wolffe1999-tx}. So, section \ref{epigen overview} provides first an overview of epigenetics phenomena and the implicated mechanisms, then section \ref{Formalism} develops the proposed attention mechanism and after,  \ref{epigenetics Model} details our Epigenetics Algorithms that exploit attention mechanisms to mimic DNA methylation. The self-supervised framework allows for the decoding phase and epigenetic operation to be learned, removing the need to encode information and improving efficiency in optimization problems. The experiments performed in the paper's final section \ref{Experiments} show that the proposed modification results in improved efficiency and a reduced risk of getting stuck in local optima by reaching the global optimum after only a few generations. 


 It encompasses a variety of mechanisms, including DNA methylation \cite{Wolffe1999-tx} (the addition of a methyl group to a DNA molecule), histone modification \cite{Bannister2011} (changes to the proteins around which DNA is wrapped), and non-coding RNA molecules, that together regulate the activity of genes.

In a sense, epigenetics acts as a sort of control system for gene expression, allowing cells to respond to changes in their environment or during development without permanently altering the underlying genetic code. These changes can be reversible and sometimes even heritable, making them a crucial factor in shaping an organism's development and evolution trajectory.

Epigenetics research has important implications for many fields, including cancer research, neuroscience, and understanding complex diseases \cite{Egger2004-gl,Feinberg1983,Simo-Riudalbas2014-xj}. In medicine, epigenetic therapies are being developed as potential treatments for cancer and other diseases by controlling gene expression or targeting specific epigenetic marks.

Overall, epigenetics provides a deeper understanding of the relationship between genes, environment, and health and offers new avenues for research and therapeutic intervention. A great example of this is identical twins, which would have identical DNA but wouldn't look the same \cite{vanDongen2021}. That's because during our life, we are going to have different experiences so that we would be different.
In the cell, many mechanisms exist for controlling how DNA is read and expressed. Several of these mechanisms, called epigenetics modification, can turn specific genes on or off without altering the DNA sequence. DNA methylation is among the best-known and widely studied mechanisms, which occurs when a methyl group is added to a DNA base. Its ability to control gene transcription dynamically inspired the proposed approach in this paper.

\section{State of the art}\label{Related work}
In recent years, epigenetics has inspired various articles in the field of computer science. In this discussion, we will examine a selection of these articles.

In \cite{Periyasamy2008}, authors propose an intragenerational epigenetic algorithm inspired by the optimization strategies used by biomolecules and present an agent-based model cell modeling and simulation environment. The authors aim to use their epigenetic algorithm to study the development patterns of cancer in different cell types that have been differentiated by trans-generational epigenetic mechanisms.

Another approach in artificial life (ALife) is presented in \cite{Sousa2010DesigningAE}, where the authors propose the EpiAL model, which uses a dynamic environment to influence the regulation of organisms and the inheritance of acquired epigenetic marks. The EpiAL model studies the plausibility of epigenetics phenomena and their relevance to an evolutionary system from an ALife perspective. The authors plan to further develop the model with a focus on biological knowledge and problem-solving techniques.

In \cite{TANEV20084469}, the authors propose a new approach to genetic programming called epigenetic programming, which incorporates explicitly controlled gene expression through histone modification. They achieve phenotypic diversity by using the cumulative effect of polyphenism and preserve individuals from the destructive effects of crossover by silencing genotypic combinations.

In \cite{Jablonka2009TransgenerationalEI}, the authors try to convince/demonstrate that epigenetic concepts can be applied to information representation and operations in computer optimization to solve various combinatorial problems. Despite the fact that they provide a pleasant introduction to the epigenetic theory, their epiGenetic Algorithm remains naive and uses random variations.

To the best of our knowledge, despite the frequent mention of the term "epigenetics", none of these studies have proposed an epigenetic algorithm that mimics well the biological mechanisms and their properties (ex., reversibility, occurring on top, ...).

In the last three to four years, AI-based models were proposed to mimic epigenetics mechanisms; in \cite{EZZARII2020}, authors propose to learn a population of epigenetics rules by training an XGBoost classifier on a given dataset. 

In a reverse context, the authors of the study\cite{Huang2021} propose A machine learning approach to brain epigenetic analysis that reveals kinases associated with Alzheimer’s disease. They show that epigenetic factors can be extracted using machine learning techniques. This motivated the development of this work mimicking epigenetics phenomena by a newly defined Deep learning framework (section \ref{epigenetics Model}) based on attention mechanism (section \ref{Formalism}).

\section{Epigenetics: an overview} \label{epigen overview}
Epigenetics is the study of changes in gene expression or cellular phenotype that do not involve alterations to the underlying DNA sequence \cite{WinNT}.  It encompasses a variety of mechanisms, including DNA methylation \cite{Wolffe1999-tx} (the addition of a methyl group to a DNA molecule), histone modification \cite{Bannister2011} (changes to the proteins around which DNA is wrapped), and non-coding RNA molecules, that together regulate the activity of genes.

In a sense, epigenetics acts as a sort of control system for gene expression, allowing cells to respond to changes in their environment or during development without permanently altering the underlying genetic code. These changes can be reversible and sometimes even heritable, making them a crucial factor in shaping an organism's development and evolution trajectory.

Epigenetics research has important implications for many fields, including cancer research, neuroscience, and understanding complex diseases \cite{Egger2004-gl,Feinberg1983,Simo-Riudalbas2014-xj}. In medicine, epigenetic therapies are being developed as potential treatments for cancer and other diseases by controlling gene expression or targeting specific epigenetic marks.

Overall, epigenetics provides a deeper understanding of the relationship between genes, environment, and health and offers new avenues for research and therapeutic intervention. A great example of this is identical twins, which would have identical DNA but wouldn't look the same \cite{vanDongen2021}. That's because during our life, we are going to have different experiences so that we would be different.
In the cell, many mechanisms exist for controlling how DNA is read and expressed. Several of these mechanisms, called epigenetics modification, can turn specific genes on or off without altering the DNA sequence. DNA methylation is among the best-known and widely studied mechanisms, which occurs when a methyl group is added to a DNA base. Its ability to control gene transcription dynamically inspired the proposed approach in this paper.

\section{Formalism} \label{Formalism}
Now we have briefly presented the epigenetics operation mimicked in this paper; we propose a new formalism implementation of the attention mechanism that fits/mimics DNA methylation well.
\subsection{Self-Reinforcement-Attention} \label{Reinforcement-Attention section}
Sometimes it is hard to deal with a traditional representation of sequence $x = [x_1, x_2, \cdots , x_p]$. It is often more useful to deal with a vectorized representation of $x$ denoted by  $v(x) = [v_i]_{i=1}^{d_v}$, where $d_v$ is the size of the vector $v(x)$.

In order to capture the relevance of each feature of the vector $v(x)$,
 we compute a representation vector $a$ named  \textbf{Reinforcement-Attention} (Figure \ref{fig:ReinforcementAttention}). Each component $a_i$ of reinforcement weights $a = [a_i]_{i=1}^{d_v}$ 
indicates the relevance of the feature $v_i$ of $v$ without any additional information. Thus, a large value $a_i$ enhances the relevance of $v_i$ while a small value of $a_i$ silences the feature $v_i$. As illustrated in figure \ref{fig:ReinforcementAttention}
the output $o = [o_i]_{i=1}^{d_v}$ is the reinforced vector according to the computed attentions $a$ defined as an element-wise multiplication of $a$ and $v$:
\begin{equation} \label{RA}
    o  = [a_i \times v_i]_{i=1}^{d_v}.
\end{equation}

{\centering
\begin{figure}
    \centering
    \includegraphics[scale=0.75]{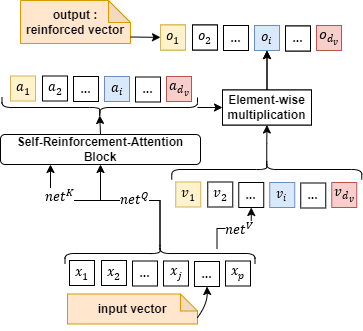}
    \caption{The Self-Reinforcement-Attention}
    \label{fig:ReinforcementAttention}
\end{figure}
}

Why Self-Reinforcement-Attention? In this sense, the elements values vector $[v_i]_{i=1}^{d_v}$ can be affected by the attention weights $[a_i]_{i=1}^{d_v}$ offering many possible regulated solutions/expressions. The attention weights vector $a$ may enhance or silence some vector's sections at strategic and specific positions that eliminate iter-elements conflicts modifying/optimizing their output function.
For example, to minimize the function :
\begin{equation} \label{example1}
f(x) = x-1,
\end{equation}
the candidate solution $x = 1.9$ represented with the values vector $v(x) = [1,9]$ of dimension $d_v=2$ fits the function better if affected by the attention weights $a^{(1)}=[0,\frac{10}{9}]$ or $a^{(2)}=[1,0]$.
The reinforcement weights $a$ may indicate that enhancing the first component/element occurs while silencing the second one leading to a candidate solution $o^{(1)} = o^{(2)} = [1,0]$.
The reinforcement weights $a$ propose alternative development pathways but expression stability to the values vector $v$.

This example corroborates research on epigenetics and specially identical twins with the same DNA sequence $x$ but two different methylations $a^{(1)}$ and $a^{(2)}$ leading to different phenotypes $o^{(1)}$ and $o^{(2)}$. Another example is to find the roots of the function:
\begin{equation} \label{example2-a}
g(x) = x^2-3x+2,
\end{equation}
which can be written as :
\begin{equation} \label{example2-b}
g(x) = (x-1.5)(x-2.5),
\end{equation}
and has two roots $x_1 = 1.5$ and $x_2 = 2.5$.

A candidate root $x=5.5$ represented with the values vector $v(x)=[5,5]$ fits the function's two roots if affected by the attention weights $a^{(1)}=[\frac{1}{5},1]$ or $a^{(2)}=[\frac{2}{5},1]$.
Once again, the reinforcement weights $a^{(1)}$ and $a^{(2)}$ propose two different development pathways allowing the candidate to reach the two roots $o^{(1)} = [1,5]$ and $o^{(2)} = [1,5]$ matching respectively $x_1 = 1.5$ and $x_2 = 2.5$. So, they confer plasticity to the candidate root. This mechanism is close to Waddington's view on epigenetics (plasticity) \cite{10.1242/jeb.120071,10.2307/2406091}; see figure \ref{fig:WaddingtonsDevelopmentalLandscapeDiagram}.


\subsection{Scaled dot-product Self-Reinforcement-Attention}
To compute attention weights $a = [a_i]_{i=1}^{d_v}$, we propose an implementation we named "scaled dot-product Reinforcement-Attention". The input consists of $d_v$ queries $Q = (q_i)_{1\le i \le d_v}$ and $d_v$ keys $K = (k_i)_{1 \le i \le d_v}$ of dimension $d_k$, and values vector $v = (v_i)_{1 \le i \le d_v}$ of dimension $d_v$ all inferred from the input vector $x$. We compute the dot products of the query $q_i$ with the key $k_i$, and divide each by $\sqrt{d_k}$ to contract an effect of large magnitude. Afterward, we apply an activation function (ex. Sigmoid,...) to obtain the reinforcement weights $A = (a_i)_{1 \le i \le d_v}$.  These steps are described by the following expression :
\begin{equation} \label{scaled_dot_produt_RA}
   a_i = f\left(\frac{q_i \cdot k_i} {\sqrt{d_k}}\right).
\end{equation}

To infer the two matrices $K$ and $Q$ and the values vector $v$, we propose three Neural networks $net^K$, $net^Q$, and $net^v$.
They could be of any architecture; the only constraint that matters is the output shapes.

As illustrated in figure \ref{fig:ScaledDotProductEgoAttention}, our implementation of reinforcement-attention is different from the traditional Scaled Dot-Product Self-Attention proposed by \cite{DBLP:journals/corr/VaswaniSPUJGKP17} as it produces a vector of dim $d_v$ instead of a matrix. 



\subsection{Multi-Head Self-Reinforcement-Attention} \label{Multi-Head Self-Reinforcement-Attention}
Instead of performing a single attention function with keys, Queries, and values, it is beneficial to perform $N$ parallel functions with different learned vectorizations of queries $Q^h$ and keys $K^h$ with $1 \le h \le N$ but keeping the same values vector $v$, yielding a series of $N$ reinforced vectors $O=[o^h]_{h=1}^{N}$.

This consideration allows one solution candidate to develop different pathways and thus explore different possibilities taking advantage of plasticity offered by the attention weights, as seen in section \ref{Reinforcement-Attention section}, equation \ref{example2-a}.

{\centering
\begin{figure}
    \centering
    \includegraphics[scale=0.75]{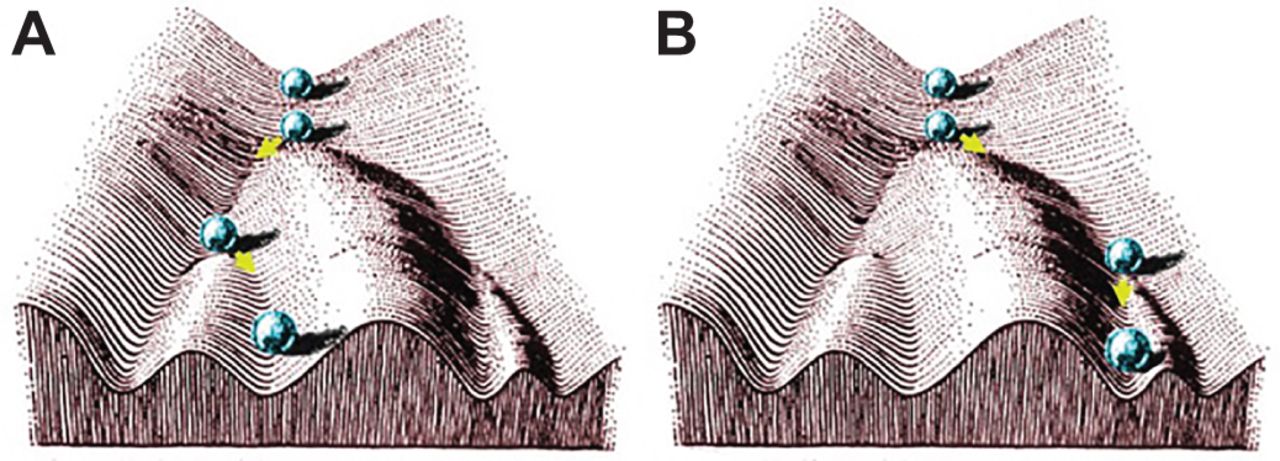}
    \caption{Waddington's developmental landscape diagram, image from :\cite{10.1242/jeb.120071}}
    \label{fig:WaddingtonsDevelopmentalLandscapeDiagram}
\end{figure}
}

{\centering
\begin{figure}
    \centering
    \includegraphics[scale=0.55]{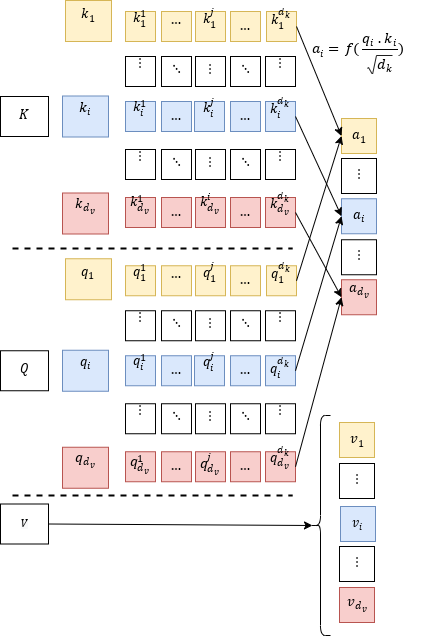}
    \caption{The Scaled Dot Product Reinforcement-Attention}
    \label{fig:ScaledDotProductEgoAttention}
    \vspace{-0.5cm}
\end{figure}
}

\section{epigenetics Model contribution}\label{epigenetics Model}
Since epigenetics is presented in the literature as a set of mechanisms and operations involved in the regulation of gene transcription, allowing the creation of phenotypic variation without genes alteration and/or chromosomes, their modeling in Artificial Life extends genetic algorithms, adding a complementary step that regulates the evaluation one. Algorithm \ref{EPiGAl} presents the epigenetics Algorithm as an extension of Genetic algorithms (lines  \ref{additionalStep1}, \ref{additionalStep2}, and \ref{additionalStep3}).

\begin{algorithm*}
\caption{The EpiGenetic Algorithm EpiGeAl}\label{EPiGAl}
\begin{algorithmic}[1]
\State \textbf{Input}: $N_i$, $p$, $r_m$, $r_c$, $DeepiGen$, $Params$. \Comment{Params = DeepGeni parameters (ex. $d_k$, $d_v$, the number of heads N, ...)}
\State\textbf{Output}: $best$, $score$, $callback$ \Comment{callback = saved copy of the DeepiGen and params that regulated the transcription of the best solution}.
\State $t\leftarrow 0$. \Comment{t = generation index}
\State $P(0) \leftarrow PopulationInitialization(N_i,p)$ \Comment{P = population}
\State $Q(0), C(0) \leftarrow \emptyset $ \Comment{Q = auxiliary populations}
\State \textbf{While not} TerminationCondition() \textbf{do}
\State\indent $DeepiGenFit(P(t), DeepiGen) $ \label{additionalStep1}
\State\indent $Scores(t) \leftarrow Evaluation(P(t),DeepiGen, Params) $ \label{additionalStep2}
\State\indent $best, score, callback = FindBest(P(t), Scores(t),DeepiGen, Params) $ \label{additionalStep3}
\State\indent $Q(t) \leftarrow Selection(P(t),Scores(t)) $ \Comment{Q contains the survivor genes}
\State\indent $C(t) \leftarrow Reproduction(Q(t)) $ \Comment{C contains the descendants genes}
\State\indent $C(t) \leftarrow Mutation(C(t),r_m) $
\State\indent $P(t+1) \leftarrow PopulationUpdating(C(t),Q(t)) $
\State\indent $t \leftarrow t+1$
\end{algorithmic}
\end{algorithm*}

\subsection{Epigenetics model stucks}
The proposed epigenetics model, DeepiGene, uses multiple blocks: a chromosome encoder and a phenotypes embedder to implement sequence to a One-real function that maps the chromosomes representation space to fitness scores $\mathbb{R}$. Figure \ref{fig:epigeneticsMechanismStack} illustrates the global architecture of the proposed DeepiGen.
Let us have a chromosome $x$ representing a  solution candidate for a given problem. It will pass throw two main blocks :  
\begin{itemize}
    \item The Chromosome Encoder Block $(CEB)$ embeds $x$ into a self-sustained representation vector of dimension $d_v$; thus, it maps the chromosomes representation space $X$ to $\mathbb{R}^{d_v}$:
        \begin{equation}
            \begin{split}
              CEB \colon X \rightarrow \mathbb{R}^{d_v} \\
              x \mapsto x_{vec}^s
            \end{split}
        \end{equation}
Hence, the chromosome $x$ is first embedded into a vector $x_{vec}$ of dimension $d_v$. This step could represent a potential solution for the encoding task. Indeed, the encoding is a problem related-task and is usually considered crucial due to its impact on the result [ref]; one does not know if his proposed encoding function fits the problem. Moreover, this offers an adaptive solution since it is learnable throw feedback.
    
Secondly, the epigenetics information that represents the self-sustaining consistency of $x$ is embedded into a vector $e_{vec}$ of the same dimension $d_v$ : 
        \begin{equation}
            \begin{split}
              EF \colon X \rightarrow \mathbb{R}^{d_v} \\
              x \mapsto e_{vec}
            \end{split}
        \end{equation}
    The self-sustained epi-chromosome $x_{vec}^s$ is the element-by-element product of $x_{vec}$ and $e_{vec}$ :
    \begin{equation}
        x_{vec}^s = e_{vec} \times x_{vec}
    \end{equation}
    
    This formulation respects the epigenetics fundamentals such as reversibility, "occurring on top", and affecting the outcome results.
    \item The phenotypes Decoder Block (PDB) takes a self-sustained epi-chromosome $x_{epi}^s$ and returns a solution $s$ that fits the optimization problem's solution space $S$: 
    \begin{equation}
            \begin{split}
              PDB \colon \mathbb{R}^{d_v} \rightarrow S\\
              x_{epi}^s \mapsto s
            \end{split}
    \end{equation}
    A fitness-evaluation step is then performed on $s$ to obtain a score.
\end{itemize}
Instead of considering a single possibility of self-sustaining consistency, we can consider different possibilities for self-sustaining consistencies generated by the epigenetics equilibrium. So for a chromosome $x$, we can get a sequence $e$ of $N$ epigenetics self-sustaining vectorizations : 
\begin{equation}
    e = [e_{vec}^h]_{1\le h \le N}
\end{equation}
with $e_{vec}^h \in \mathbb{R}^{d_v}$. The Chromosome Encoder Block (CEB) will return a sequence of $N$ self-sustained epi-chromosomes $[x_{vec}^{h,s}]_{1\le h \le N}$ satisfying :
        \begin{equation}
            \begin{split}
              CEB_M \colon X \rightarrow \mathbb{R}^{d_v \times N} \\
              x \mapsto [x_{vec}^{h,s}]_{1\le h \le N}
            \end{split}
        \end{equation}
with $x_{vec}^{h,s} = e_{vec}^h \times x_{vec}$. Then, the Phenotypes Decoder Block (PDB) will return $N$ potential solutions $[s^h]_{1\le h \le N}$ corresponding to the $N$ epi-chromosomes $[x_{vec}^{h,s}]_{1\le h \le N}$. 
Finally, we propose to apply optimum pooling after evaluating the fitness of each solution.

This consideration allows different expressions $s = \{s^1,\cdots,s^h,\cdots,s^N\}$ for the chromosome $x$. This last can take different permitted trajectories, leading to different outcomes or solutions that fit local constraints. 

{\centering
\begin{figure}
    \centering
    \includegraphics[scale=0.75]{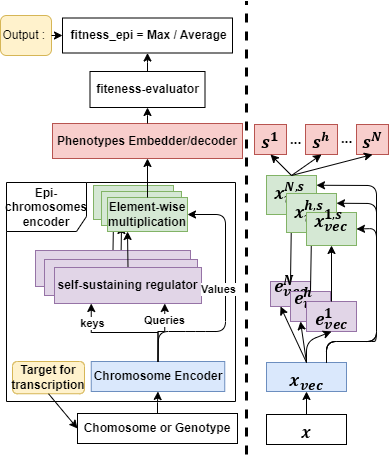}
    \caption{The proposed architecture to consider epigenetics as self-sustaining regulation operators}
    \label{fig:epigeneticsMechanismStack}
    \vspace{-1em}
\end{figure}
}

\subsection{Application of Self-Reinforcement-Attention in our epigenetics Model}
As mentioned, Our DeepiGen uses the proposed self-reinforcement-attention in the chromosome Encoder Block (CEB).
Indeed, the self-sustaining consistency $e_{vec}$ can be seen as the reinforcement weights $a$ (Cf. \ref{Reinforcement-Attention section}). Then, it can be embedded using the scaled-dot-product Self-Reinforcement-attention. Thus, the self-sustained epi-chromosomes $[x_{vec}^{h,s}]_{1\le h \le N}$ correspond to the reinforced vectors in $O=[o^h]_{h=1}^{N}$.

The reinforcement-attention mechanism tries to catch incompatibilities between genes in the chromosomes and gives a way to regulate the encoded information in the chromosome, mimicking the self-regulation process in biology.
\subsection{Training strategy for epigenetics model} 
We chose to train our epigenetics model DeepiGen to optimize the objective function, i.e., to maximize the population scores. It consists in training the neural network with the loss function :
\begin{equation}
\mathcal{L} = \sum_{n=1}^{N_i} \frac{1}{score[n]}
\end{equation}

In the case of multi-head-reinforcement attention, we add a diffusion constraint to force the inter-head dispersion. The occurrence of such a phenomenon will be controlled with a diffusion rate $r_{diff}$.  Thus, we will maximize the variance for all transcriptions associated with each chromosome of the population at some generations.
\subsection{Framework properties}\label{Framework properties}
Several properties flow from our approach:
\begin{itemize}
    \item A smart chromosome encoding: The proposed approach automates the decoding phase via the block (PDB). Furthermore, with learnable encoding using a neural network, the proposed algorithm provides an excellent answer to the old classic question: how to encode chromosomes? Thus, we can use binary codes for all problems.
    \item An equilibrium between exploration and exploitation: classic GAs require clever exploration through the mutation rate; a high value for the mutation rate leads to an unstable system that does not exploit the explored region well. In our proposed algorithm, the neural network DeepiGen focuses on exploiting the new potential area, which could make a Nash equilibrium reachable.
\end{itemize}

\section{Experiments} \label{Experiments}
One of the limitations of Genetic algorithms is that they tend to converge towards local optima or even arbitrary points rather than the global optimum for non-convex problems. To investigate the behavior of our approach, we decided to focus on multi-peaks optimization problems.

\subsection{Optimization of multi-peaks problems: studies in convergences and dynamics}
To study the convergence of the proposed epigenetics framework and dynamic on non-convex problems, we propose testing it on two well-known multi-peaks optimization problems: BUMPY \cite{KEANE199575}, and STALAGMITE functions \cite{Stalagmite:2020}.

\subsubsection{The BUMPY function} \cite{KEANE199575} is one of the multi-peak optimization problems that produce a series of peaks getting smaller with distance from the origin. it's objective function can be defined for values $(x,y)$ in $\mathcal{D}_B = ]0,10]^2$ as :
$$f(x,y) = \frac{sin^2(x-y) sin^2(x+y)}{\sqrt{x^2+2y^2}}$$
The BUMPY problem is dependent on only two variables $x$ and $y$, which aids the display of the function and the qualitative discussion of DeepiGen results in section \ref{Results}. Figure \ref{fig:3DBumpy_func_0_10} shows the true maximum $0.675$ reached at $(1.393,0.006)$ but exhibits some additional properties of the given problem that make it kind of tricky to solve: 
\begin{itemize}
    \item First, the problem is quite symmetrical in $x=y$, so a similar peak of height $0.47$ at $(0.031,1.441)$, which could be quite competitive if the algorithm gets trapped in this local optimum.
    \item The next two peaks are $0.365$ and $0.274$ at $(1.593,0.471)$ and $(0.475,1.578)$ respectively. These two peaks are less competitive than the first.
\end{itemize}
Thus, the first challenge was to test our approach's ability to reach the global optimum of height $0.675$ at $(1.393,0.006)$, even if the initial population was designed to intentionally favor the local optimum of height $0.47$ at $(0.031,1.441)$, see figure \ref{fig:initial_population_1}.

The second challenge for the BUMPY function is to evaluate the impact of considering the multi-heads architecture \ref{Multi-Head Self-Reinforcement-Attention}. i.e., we are interested in the question: what do other heads do? Do they explore solutions other than the global one, or will they only collapse into the same optimum?

{\centering
\begin{figure}
    \centering
    \includegraphics[scale=0.45]{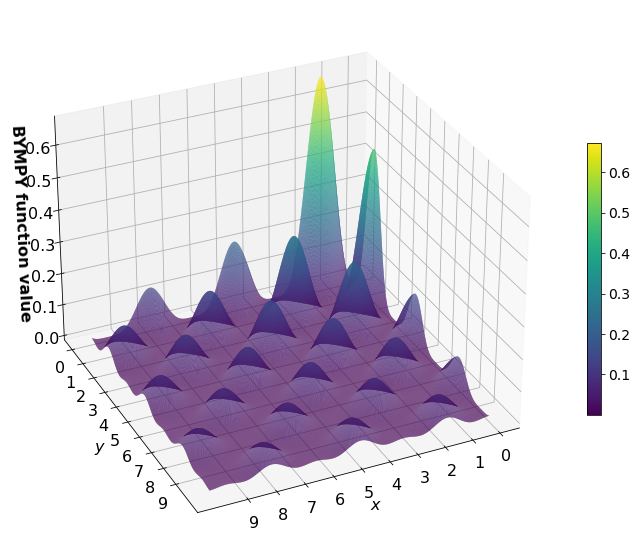}
    \caption{The Surface generated by BUMPY function for $0 < x,y \leq 10$}
    \label{fig:3DBumpy_func_0_10}
\end{figure}
}

\subsubsection{The STALAGMITE function} \cite{Stalagmite:2020} introduced as a multi-peaks optimization problem for SKILL LYNC's students in 2020. It was inspired by stalagmite structures formed on some caves floors and can be defined on the domain $\mathcal{D}_S = [0,1]^2$ as a product of four functions :
$$f(x,y) = f_{1,x}f_{2,x}f_{1,y}f_{2,y}.$$
where,
$$f_{1,x} = [sin(5.1\pi x + 0.5)]^6,$$
$$f_{1,y} = [sin(5.1\pi y + 0.5)]^6,$$
$$f_{2,x} = exp[-4ln(2)\frac{(x-0.0667)^2}{0.64}],$$
$$f_{2,y} = exp[-4ln(2)\frac{(y-0.0667)^2}{0.64}],$$
as one can see, the STALAGMITE function is symmetrical i.e., 
$$\forall (x,y) \in \mathcal{D}_S, f(x,y) = f(y,x).$$ 

Figure \ref{fig:3DStalagmite_func_0_1} shows the global maximum of height $1$ reached at $(0.06,0.06)$ but exhibits another property of the stalagmite problem: two additional peaks of height $0.845$ at $(0.262,0.067)$ and $(0.067,0.262)$, which could be pretty competitive if the algorithm gets trapped in one of these local optima.

So, this property raises the question: what would be the behavior of our algorithm if a mathematical constraint forces ignore the global optimum of height $1$ at $(0.06,0.06)$? Since two local optima are possible, will the algorithm be sensitive to the population initialization for finding the new best global optimum (the old local optima $0.845$ at $(0.262,0.067)$ or $(0.067,0.262)$)?

{\centering
\begin{figure}
    \centering
    \includegraphics[scale=0.45]{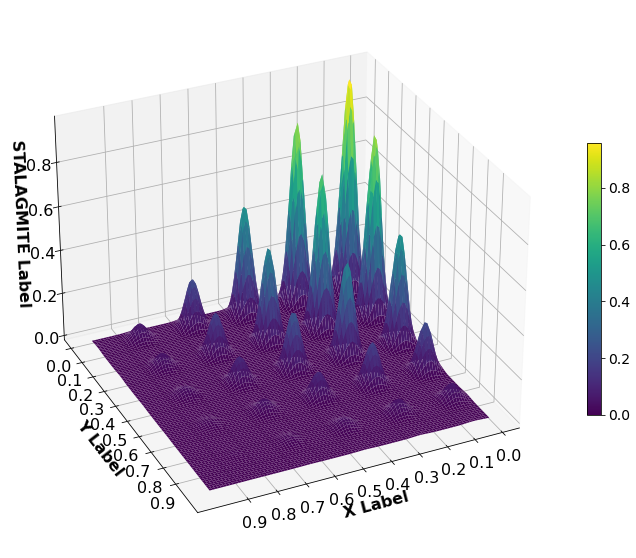}
    \caption{The Surface generated by STALAGMITE function for $0\leq x,y \leq 1$}
    \label{fig:3DStalagmite_func_0_1}
\end{figure}
}

{\centering
\begin{figure}
    \centering
    \includegraphics[scale=0.45]{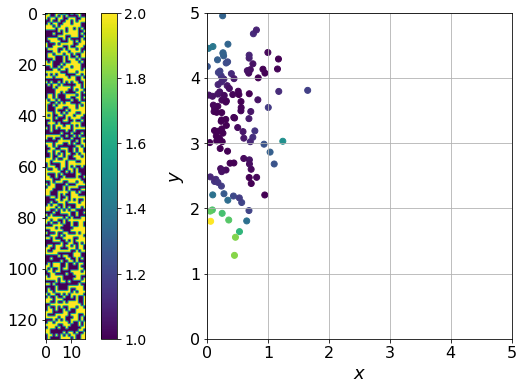}
    \caption{The initial population considered for experiment }
    \label{fig:initial_population_1}
\end{figure}
}

{\centering
\begin{figure}
    \centering
    \includegraphics[scale=0.45]{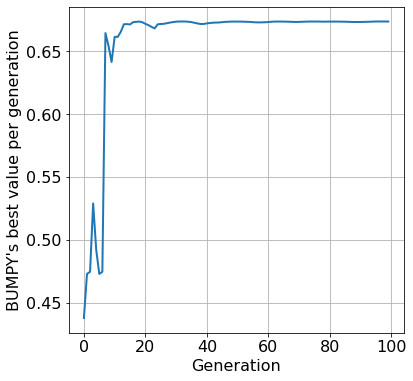}
    \caption{evolution of the best-found value per generation for the BUMPY function}
    \label{fig:best_per_generation_1}
\end{figure}
}

{\centering
\begin{figure*}
    \centering
    \includegraphics[scale=0.55]{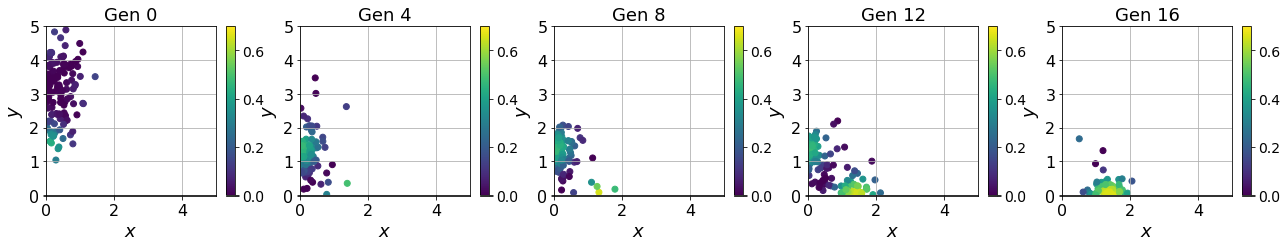}
    \caption{The population migration over generations to reach the optimal global peak of height $0.675$ for the BUMPY function at position $(1.39,0.006)$}
    \label{fig:pop_per_generation_1}
\end{figure*}
}

\subsection{Experiment outline} \label{Experiment outline}
To ensure that our approach provides good answers for the four challenges announced above, four experiments were conducted : 
\begin{itemize}
    \item Experiment 1 (the global optimum): we tested the ability of our approach to reach the global optimum with the basic model. i.e., one Self-Reinforcement-Attention head. For this, different configurations were tested for Benchmarking on both BUMPY and STALAGMITE functions. A particular interest was initially given to the population to favor the local optimum intentionally.
    \item Experiment 2 (twins): we tested the impact of considering multi-heads Self-Reinforcement-attention architecture. In other words, will the algorithm be able to find more than one optimal solution? i.e., will the different heads focus on different regions and/or characteristics?
\end{itemize}



\subsection{Parametrization}
Our DeepiGen has ten degrees of freedom: first, the five well-known classical parameters of Genetic algorithms :
\begin{enumerate}
    \item $N_i$ : the population size.
    \item $p$ the dimension of chromosomes.    
    \item $r_{mut}$ : the mutation rate.
    \item $r_c$ the crossover rate.
    \item $nb_{iter}$ the maximum number of iterations.
\end{enumerate}
In addition to the five parameters of our proposed model DeepGeni:
\begin{enumerate}
    \setcounter{enumi}{5}
    \item $d_k$ : the output dimension of the keys and Queries encoders.
    \item $embed$ : a Boolean value to determine if the chromosomes will be embedded.
    \item $d_v$ : the output dimension of the values embedder (N.B.$d_v=p$ and $net^V$ is Identity transfomation if $embed==False$).
    \item $N$: the number of heads in the DeepGeni model.
    \item $r_{diff}$ : the diffusion rate that control the inter-heads dispersion.
\end{enumerate}

For the two main experiences announced in \ref{Experiment outline}, one thousand (1000) different configurations, varying these parameters, were tested to ensure the robustness of the proposed algorithm.
Thus, for classical GAs parameters: the population size parameter $N_i$ took values in $[32,64,128,256,512,1024]$, the dimension of chromosomes $p$ took values in $[8,16,32]$, the mutation rate $r_{mut}$ varies between $0.05$ and $0.3$ with a step of $0.01$, the crossover rate $r_c$ took values in $[0.7,0.8,0.9]$ allowing some chromosomes to survive, and finally the maximum number of iterations was set to $nb_{iter}=100$.\\
Furthermore, for the DeepiGen parameters: the dimension of Queries and keys dimension $d_k$ took values in $[16,32,64]$, the values dimension $d_v$ took values in $[8,16,32,64]$ for all cases of $embed==True$. The important parameter was the number of Self-reinforcement-Attention heads $N$, which took values in $[1,8,16]$. So, the diffusion rate $r_{diff}$ took values between $0.05$ and $0.3$ with a step of $0.05$ for all $N>1$. \\
\indent However, as discussed in section \ref{Framework properties}, only binary encoding was considered in this paper. The real encoding was assumed to be implicitly realized by the value encoder $net^v$.
The last thing to test was the impact of initialization for Population on the obtained results, so difficult scenarios were considered intentionally; for example, the initial population considered respectively for both BUMPY and STALAGMITE functions favored local optima.

\subsection{Results and discussion}\label{Results}
\subsubsection{Experiment 1} over the one thousand (1000) tested configurations, all succeed in reaching/finding the global optimum for both BUMPY and STALAGMITE functions. Furthermore, considering a constraint that penalizes the global optimum, our algorithm found the next best optimum easily. \\
In this section, we chose to present the obtained results for the BUMPY function considering the following configuration: an initialized population of $N_i=128$ chromosomes of dimension $p=16$, a mutation rate of value $r_{mut} = 0.1$, and crossover rate of $r_c=0.7$, for Keys and Queries output dimension the value of $d_k=32$ is considered. Moreover, no value embedding, hence $embed=False$, and only one head $N=1$. \\
\indent Figure \ref{fig:initial_population_1} shows the initial population of $N_i = 128$ chromosomes at left and the corresponding decoded values $(x,y)$ at right. It shows that all of them are near the second peak of height $0.47$ in position $(0.031,1.441)$, so all convex optimization methods (ex., gradient descent) risk getting stuck with this local optimum.
Figure \ref{fig:best_per_generation_1} shows the evolution of the best optimum value over generations. One can see that our algorithm reached the global optimum at the sixteenth generation $t = 16$. Actually, the algorithm began considering this convex area of the global optimum at the eighth generation $t=8$.\\
\indent Q: What did happen during the training?

As expected and can be seen in figure \ref{fig:pop_per_generation_1}, the local optimum was reached during the first generation due to the optimization step (7). At the fourth generation $t=4$, the model began exploring the global convex area; at the eighth generation $(t= 8)$, the decoded solution matches a solution near the global optimum $s_{opt}=0.67$. After this discovery, the model focused on this optimal convexity around this global optimum for the next generations. thus, the inertia decreases, and all solutions collapse into the global optimum.\\
It is worth noting that sometimes mutations occur and try to explore regions far from the centered values.
\subsubsection{Experiment 2}
considering multiple heads for paying attention and hence enhancing/silencing some genes allows the algorithm to consider multiple optimal values associated with the same chromosome as announced in section \ref{Reinforcement-Attention section} for the BUMPY function case. Of course, depending on the area considered by the attention head. 

Indeed, with a number of heads $N=16$ and a diffusion rate of value $r_{diff}=0.3$, our algorithm mimicked perfectly the epigenetic observation noticed for identical twins; to the same chromosome, two different heads succeeded in associating/reaching two different optimal peaks.

Figure \ref{fig:identical_twins} shows the evolution of the best optimal values over generations for the BUMPY function; we can see that different heads collapsed into three main clusters around the two best and second best optimal peaks at generation $t=36$ and continue to explore other regions as shown at generation $t=48$.
Thus, given the binary encoded chromosome: 
$$
x = [0.0, 1.0, 0.0, 0.0, 0.0, 0.0, 0.0, 0.0, 1.0, 0.0, 0.0, 0.0, 1.0, 0.0, 1.0, 0.0] $$
The Chromosome Encoder Block $(CEB_M)$ with multiple \\Self-Reinforcement-Attention heads, returns several Self-sustained epi-chromosomes (Self-Reinforced vectors) $[x_{vec}^{h,s}]_{1\leq h \leq 16}$. After training, the two heads $h= 9$ and $h=15$ of our $(CEB_M)$ propose for the given chromosome $x$, the two Self-sustained epi-chromosomes:

\begin{align*}
x_{vec}^{9,s}=[&0.89, 1.06, 0.41, 0.74, 0.7,  0.71, 0.78, 0.43,\\ &0.45, 1.3,  0.77, 0.46, 0.84, 1.73, 1., 0.51],    
\end{align*}

 and
 \begin{align*}
 x_{vec}^{15,s} = [&0.66, 0.34, 1.74, 0.56, 0.68, 1.22, 0.43, 0.46,\\ &1.37, 1.26, 0.42, 0.78, 0.68, 0.64, 0.53, 1.25],
 \end{align*}
illustrated in figure \ref{fig:identical_twins_choromo_activation} to what the trained Phenotypes Decoder Block $(PDB)$ returns the two decoded chromosomes: 
$$PDB(x_{vec}^{9,s}) = (1.426,0.036)$$
and
$$PDB(x_{vec}^{15,s}) = (0.166,1.354)$$
As we can see in figure \ref{fig:identical_twins_choromo_activation} and the values $x_{vec}^{9,s}$ and $x_{vec}^{15,s}$, to activate the first solution $(1.426,0.036)$ the gene at position $13$ was enhanced while genes at positions $2$, $7$ and $8$ were silenced. Similarly, to activate the second solution $(0.166,1.354)$ the gene at position $2$ was enhanced while genes at positions $6$, $7$, $10$, and $14$ were silenced.
On the other hand, considering the STALAGMITE function with a constraint that forces ignoring the global optimum succeeded in reaching the two symmetrical solutions of height $0.845$ at positions $(0.265,0.065)$ and $(0.065,0.265)$ but with two different chromosomes through two different heads $h=5$ and $h=13$.

Results obtained in Experiment 2 corroborate the Waddington's developmental landscape diagram shown in figure \ref{fig:WaddingtonsDevelopmentalLandscapeDiagram}.
{\centering
\begin{figure*}
    \centering
    \includegraphics[scale=0.55]{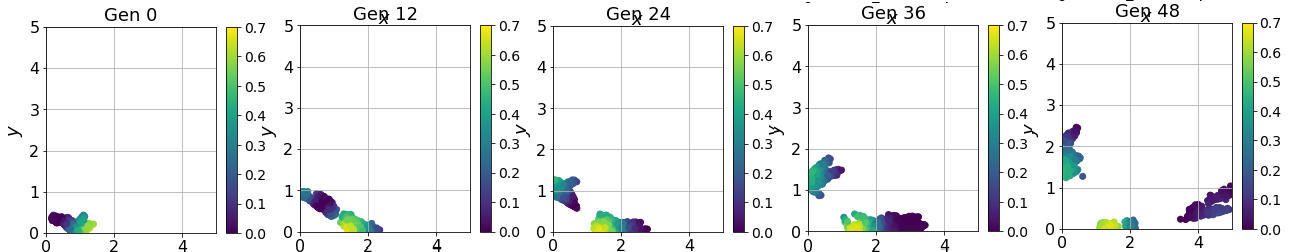}
    \caption{The population migration over generations to reach the multi-optimal peaks of height $0.675$ and $0.475$ respectively for the BUMPY function at positions $(1.39,0.006)$ and $(0.365,0.274)$ respectively}
    \label{fig:identical_twins}
\end{figure*}
}

{\centering
\begin{figure*}
    \centering
    \includegraphics[scale=0.55]{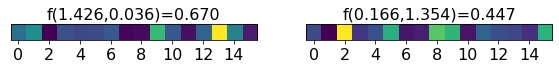}
    \caption{The two attention arrays proposed by respectively head $8$ and $15$ for the same chromosome to reach an approximating global optimum $0.67$ and an approximating second best optimum $0.44$ respectively for the BUMPY function at positions $(1.426,0.036)$ and $(0.166,1.354)$ respectively}
    \label{fig:identical_twins_choromo_activation}
\end{figure*}
}




\section{Conclusion and current work}
This paper presented new Epigenetics algorithms: a modification of the genetic algorithms (GAs) using a novel attention mechanism. The modification mimics the epigenetic process of DNA methylation (enhancing/silencing). One of the main contributions is that our approach is self-supervised; The decoding phase and epigenetics operation are learnable. This motivates dropping age-old questions of how to encode information. 
Later, and after discussing similarities between Attention and epigenetics mechanisms, the experiments in the final section showed that the proposed modification allowed the algorithm to fit the multi-peaks optimization problems efficiently. The efficiency is improved by reaching the global optimum after a few generations and thus reducing the risk of getting stuck in local optima. Besides, using multiple attention heads showed that the same chromosome can represent several optimums depending on the region. 

Furthermore, the presented results are part of the research on the possibility of mimicking epigenetic processes in genetic algorithms using attention mechanisms. As part of our research, it is planned to study the convergence of our approach theoretically and also examine the assembly of other modifications that mimic other epigenetic processes in various genetic algorithms; thus, research is also being carried out into the development of an epigenetic algorithm that considers external factors.





\newpage

\bibliographystyle{apalike} 
\bibliography{referencias.bib} 

\end{document}